\documentclass[sigconf,natbib=false]{acmart}

\usepackage[style=ACM-Reference-Format,backend=bibtex,sorting=none]{biblatex}
\usepackage[acronyms]{glossaries}
\usepackage{subcaption}
\usepackage{float}
\usepackage{booktabs}
\usepackage{tabularx}
\usepackage{csquotes}
\usepackage[english]{babel}
\usepackage{balance}
\addto\extrasenglish{

}
\AtBeginDocument{%
  \providecommand\BibTeX{{%
    \normalfont B\kern-0.5em{\scshape i\kern-0.25em b}\kern-0.8em\TeX}}}

\setcopyright{acmcopyright}
\copyrightyear{2018}
\acmYear{2018}
\acmDOI{XXXXXXX.XXXXXXX}

\acmConference[Conference acronym 'XX]{Make sure to enter the correct
  conference title from your rights confirmation emai}{June 03--05,
  2018}{Woodstock, NY}
\acmPrice{15.00}
\acmISBN{978-1-4503-XXXX-X/18/06}

\newacronym{htm}{HTM}{Hierarchical Temporal Memory}
\newacronym{tm}{TM}{Temporal Memory}
\newacronym{sp}{SP}{Spatial Pooler}
\newacronym{sdr}{SDR}{Sparse Distributed Representation}
\newacronym{mlp}{MLP}{Multilayer Perceptron}
\newacronym{relu}{ReLU}{Rectifying Linear Unit}
\newacronym{sgd}{SGD}{Stochastic Gradient Descent}
\newacronym{mse}{MSE}{Mean Squared Error}
\newacronym{cnn}{CNN}{Convolutional Neural Network}
\newacronym{bn}{BN}{Batch Normalization}
\newacronym{rnn}{RNN}{Recurrent Neural Network}
\newacronym{gan}{GAN}{Generative Adversarial Network}
\newacronym{ae}{AE}{Autoencoder}
\newacronym{vae}{VAE}{Variational Autoencoder}
\newacronym{orb}{ORB}{Oriented FAST and Rotated BRIEF}

\begin{document}

\title{Grid HTM: Hierarchical Temporal Memory for Anomaly Detection in Videos}

\author{Vladimir Monakhov}
\affiliation{
  \institution{University of Oslo and SimulaMet}
 \country{Norway}
}
\author{Vajira Thambawita }
\affiliation{%
  \institution{SimulaMet}
  \country{Norway}
}
\author{P\aa l Halvorsen}
\affiliation{%
  \institution{SimulaMet and OsloMet}
  \country{Norway}
}
\author{Michael A. Riegler}
\affiliation{%
  \institution{SimulaMet and UiT}
   \country{Norway}
}
\renewcommand{\shortauthors}{Monakhov et al.} 

\begin{abstract}
The interest for video anomaly detection systems has gained traction for the past few years. The current approaches use deep learning to perform anomaly detection in videos, but this approach has multiple problems. For starters, deep learning in general has issues with noise, concept drift, explainability, and training data volumes. Additionally, anomaly detection in itself is a complex task and faces challenges such as unknowness, heterogeneity, and class imbalance. Anomaly detection using deep learning is therefore mainly constrained to generative models such as generative adversarial networks and autoencoders due to their unsupervised nature, but even they suffer from general deep learning issues and are hard to train properly. In this paper, we explore the capabilities of the Hierarchical Temporal Memory (HTM) algorithm to perform anomaly detection in videos, as it has favorable properties such as noise tolerance and online learning which combats concept drift.  We introduce a novel version of HTM, namely, Grid HTM, which is an HTM-based architecture specifically for anomaly detection in complex videos such as surveillance footage.
\end{abstract}



%
\maketitle

\section{Introduction}
As the global demand for security and automation increases, many seek to use video anomaly detection systems. In the US alone, the surveillance market is expected to reach $\$23.60$ Billion by 2027~\cite{us_video_stats}. Leveraging modern computer vision, modern anomaly detection systems play an important role in increasing monitoring efficiency and reducing the need for expensive live monitoring. Their use cases can vary from detecting faulty products on an assembly line to detecting car accidents on a highway.
\par
The most important component in video anomaly detection systems is the intelligence behind it. The intelligence ranges from simple on-board algorithms to advanced deep learning models, where the latter has experienced increased popularity in the past few years.
Yet, despite the major progress within the field of deep learning, there are still many tasks where humans outperform models, especially in anomaly detection where the anomalies are often undefined. Deep learning approaches also perform poorly when dealing with noise and concept drift.
\par
The cause for the discrepancy lies in the difference between how humans and machine learning algorithms represent data and learn. Most machine learning algorithms use a dense representation of the data and apply back-propagation in order to learn. Human learning happens in the neocortex, where evidence points to that the neocortex uses a sparse representation and performs Hebbian-style learning. For the latter, there is a growing field of machine learning dedicated to replicating the inner mechanics of the neocortex, namely  \gls*{htm} theory~\cite{BAMI}. This theory outlines its advantages over standard machine learning, such as noise-tolerance and the ability to adapt to changing data.
\par

With the advantages of  \gls*{htm} and the rise of video anomaly detection in mind, a natural question one could pose is whether it is possible to apply  \gls*{htm} for anomaly detection in videos. Combined with a lack of related works, it is this very question that is the motivation behind this paper. In this paper, we therefore propose and evaluate Grid HTM which is a novel expansion of the base HTM algorithm that allows for unsupervised anomaly detection in videos.

\section{Background}

Anomaly detection is often defined as detecting data points that deviate from the general distribution~\cite{anomaly_detection}. Unlike most other problems in deep learning, anomaly detection deals with unpredictable and rare events which makes it hard to apply traditional deep learning for anomaly detection. 
A subset of anomaly detection is smart surveillance~\cite{anomalyvideo}, which is the use of video analysis specifically in surveillance. 
\par
An issue for deep-learning models in general is that they are susceptible to noise in the dataset~\cite{noise1,noise2}, which leads to decreased model accuracy and poor prediction results. Due to the nature of training deep learning models, they are also in most cases not self-supervised and therefore require constant tuning in order to stay effective on changing data. In addition, they require a lot of data before they can be considered effective, and performance increases logarithmically based on the volume of training data~\cite{deeplearning_dataset}. Deep learning models also suffer from issues with out-of-distribution generalization~\cite{deeplearning_ood_generalization_survey}, where a model might perform great on the dataset it is tested on, but performs poorly when deployed in real life. This could be caused by selection bias in the dataset or when there are differences in the causal structure between the training domain and the deployment domain~\cite{deeplearning_ood}. 
Another challenge with deep learning models is that they generally suffer from a lack of explainability~\cite{XAI}. While it is known \emph{how} the models make their decisions, their huge parametric spaces make it unfeasible to know \emph{why} they make those predictions. Combined with the vast potential that deep learning offers in critical sectors such as medicine, makes approaches that offer explainability highly attractive.
\par
The HTM theory~\cite{BAMI} introduces a machine learning algorithm which works on the same principles as the brain and therefore solves some of the issues that deep learning has. \gls*{htm} is considered noise resistant and can perform online learning, meaning that it learns as it observes more data. \gls*{htm} replicates the structure of the neocortex which is made up of cortical regions, which in turn are made up of mini-columns and then neurons.

The data in an \gls*{htm} model is represented using a \gls*{sdr}, which is a sparse bit array. An encoder converts real world values into \glspl*{sdr}, and there are currently encoders for numbers, geospatial locations, categories, and dates. 
One of the difficulties with \gls*{htm} is making it work on visual data, where creating a good encoder for visual data is still being researched~\cite{CNN_HTM, eyeencoder, MotionAnomalyDetection}.
The learning mechanism consists of two parts, the \gls*{sp} and the \gls*{tm}. The \gls*{sp} learns to extract semantically important information into output \glspl*{sdr}. The \gls*{tm} learns sequences of patterns of \glspl*{sdr} and forms a prediction in the form of a predictive \gls*{sdr}. 
A research study~\cite{AHMAD2017134} has shown that \gls*{htm} is very capable of performing anomaly detection on low-dimensional data and is able to outperform other anomaly detection methods. However, related works, such as \textcite{MotionAnomalyDetection}, show that \gls*{htm} struggles with higher dimensional data. Therefore, a natural conclusion is that HTM should be applied differently, and that a new type of architecture using HTM should be explored for the purpose of video anomaly detection and surveillance.


\section{Grid HTM}
\begin{figure}[htb]
    \centering
    \begin{subfigure}[t]{0.5\linewidth}
        \centering
        \includegraphics[width=\linewidth]{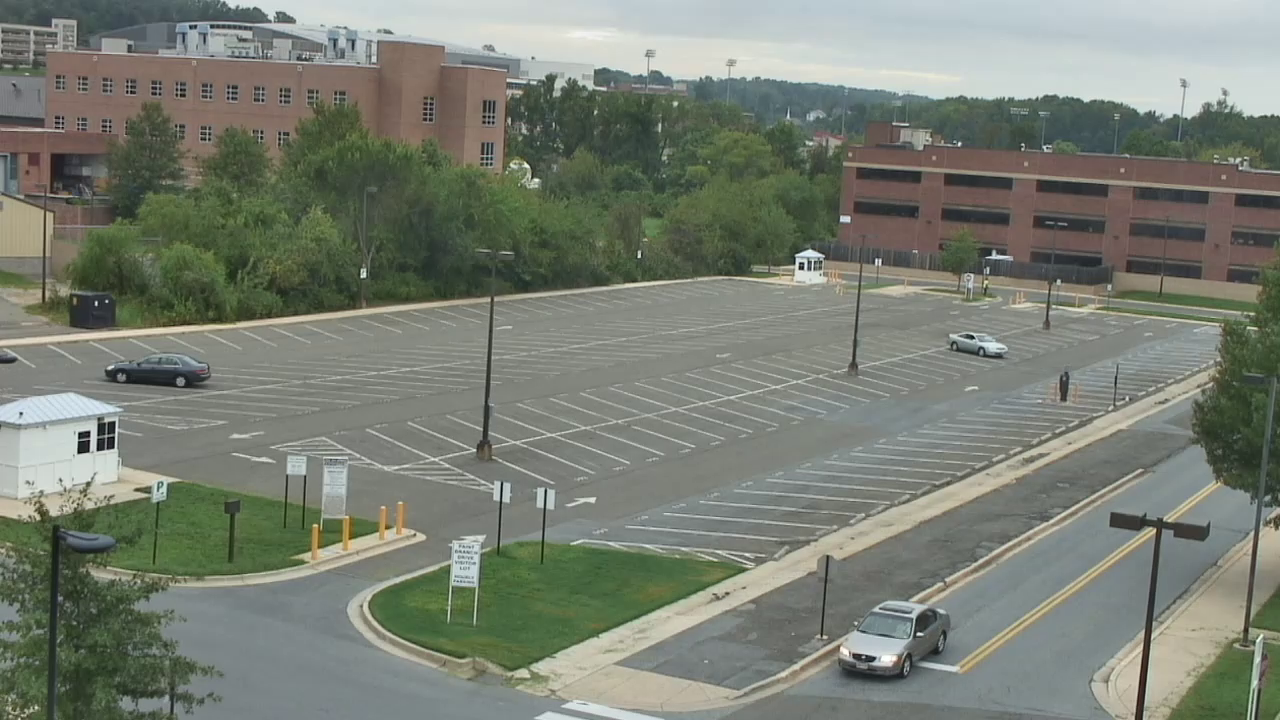}

    \end{subfigure}%
    \begin{subfigure}[t]{0.5\linewidth}
        \centering
        \includegraphics[width=\linewidth]{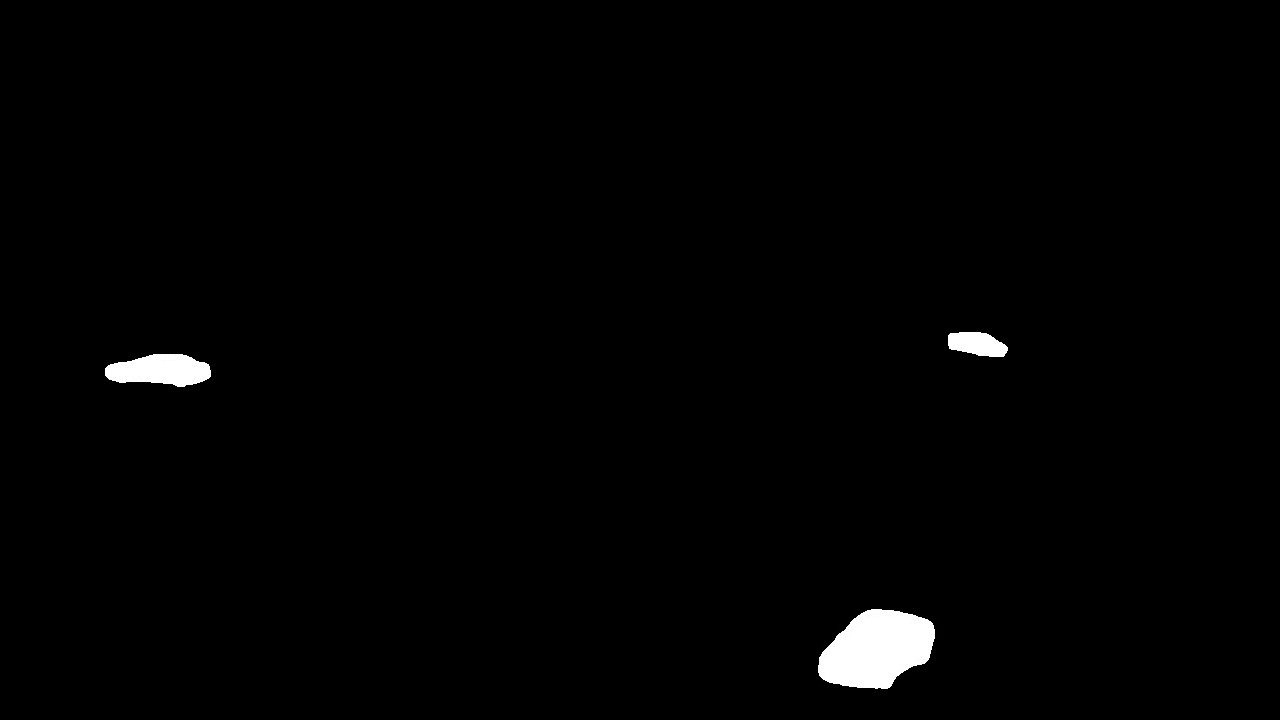}
    \end{subfigure}
    \caption[Segmentation Result of Cars]{Segmentation result of cars, which is suited to be used as an SDR. Original frame taken from VIRAT~\cite{VIRAT}.}
    \label{fig:example_segmentation}
\end{figure}

This paper proposes and explores a new type of architecture, named Grid HTM, for anomaly detection in videos using \gls*{htm}, and proposes to use segmentation techniques to simplify the data into an SDR-friendly format. These segmentation techniques could be anything from simple binary thresholding to deep learning instance segmentation. Even keypoint detectors such as \gls*{orb}~\cite{orb_detector} could in theory be applied. When explaining Grid HTM, the examples will be taken from deep learning instance segmentation of cars on a video from the VIRAT~\cite{VIRAT} dataset. An example segmentation is shown in \autoref{fig:example_segmentation}.


\begin{figure}[htb]
    \centering
    \begin{subfigure}[t]{0.49\linewidth}
        \centering
        \includegraphics[width=\linewidth]{methodology/car_segmentation.png}

    \end{subfigure}%
    \hspace{0.5mm}
    \begin{subfigure}[t]{0.49\linewidth}
        \centering
        \includegraphics[width=\linewidth]{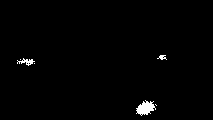}
    \end{subfigure}
    \caption[SDR and SP Representation]{The SDR (left) and its corresponding \gls*{sp} representation (right). Note that the \gls*{sp} is untrained.}
    \label{fig:car_segmentation_sp}
\end{figure}

The idea is that the \gls*{sp} will learn to find an optimal general representation of cars. How general this representation is can be configured using the various \gls*{sp} parameters, but ideally they should be set so that different cars will be represented similarly while trucks and motorcycles will be represented differently. An example representation by the SP is shown in \autoref{fig:car_segmentation_sp}.


\par
The task of the \gls*{tm} will then be to learn the common patterns that the cars exhibit, their speed, shape, and positioning will be taken into account. Finally, the learning will be set so that new patterns are learned quickly, but forgotten slowly. This will allow the model to quickly learn the norm, even if there is little activity, while still reacting to anomalies. This requires that the input is stationary, in our example this means that the camera is not moving.
\par
It is possible to split different segmentation classes into their respective SDRs. This will give the SP and the TM the ability to learn different things for each of the classes. For instance, if there are two classes "person" and "car", then the TM will learn that it is normal for objects belonging to "person" to be on the sidewalk, while objects belonging to "car" will be marked as anomalous when on the sidewalk.
\par
Ideally, the architecture will have a calibration period spanning several days or weeks, during which the architecture is not performing any anomaly detection, but is just learning the patterns.
\section{Improvements}
\textcite{MotionAnomalyDetection} tested only the base HTM version and showed that the algorithm cannot handle subtle anomalies, therefore multiple improvements needed to be introduced to increase effectiveness.

\textbf{Invariance. }One issue that becomes evident is the lack of invariance, due to the \gls*{tm} learning the global patterns. Using the example, it learns that it is normal for cars to drive along the road but only in the context of there being cars parked in the parking lot. It is instead desired that the \gls*{tm} learns that it is normal for cars to drive along the road, regardless of whether there are cars in the parking lot.
We proposes a solution based on dividing the encoder output into a grid and have a separate \gls*{sp} and \gls*{tm} for each cell in the grid. The anomaly scores of all the cells are then aggregated into a single anomaly score using an aggregation function.

\textbf{Aggregation Function. } Selecting the correct aggregation function is important because it affects the final anomaly output. For instance, it might be tempting to use the mean of all the anomaly scores as the aggregation function:
\begin{align*}
    X         & :\{x \in \mathbb{R} : x \geq 0\}
    \\
    Anomaly\_Score & =\dfrac{\sum\limits_{x \in X}x}{|X|}
\end{align*}
Where $X$ denotes the set of anomaly scores $x$ from each individual grid.
However, this leads to problems with normalization, meaning that an overall anomaly score of 1 is hard to achieve due to many cells having a zero anomaly score. In fact, it becomes unclear what a high anomaly score is anymore. Using the mean also means that anomalies that take up a lot of space will be weighted higher than anomalies that take up a little space, which might not be desirable.
To solve the aforementioned problem and if the data has little noise, a potential aggregation function could be the non-zero mean:
\begin{align*}
    X         & :\{x \in \mathbb{R} : x > 0\}
    \\
    Anomaly\_Score & =
    \begin{cases}
        \dfrac{\sum\limits_{x \in X}x}{|X|} & \text{if } |X| > 0 
        \\
        \hfil 0                             & \text{otherwise}
    \end{cases}
\end{align*}
Meaning that only the cells with a strictly positive anomaly score, will be contributing to the overall anomaly score which helps solve the aforementioned normalization and weighting problem. On the other hand, the non-zero mean will perform poorly when the architecture is exposed to noisy data which could lead to there always being one or more cells with a high anomaly score.
\begin{figure}[htb]
    \centering
    Noisy data
    \begin{subfigure}[t]{0.5\linewidth}
        \centering
        \includegraphics[width=\linewidth]{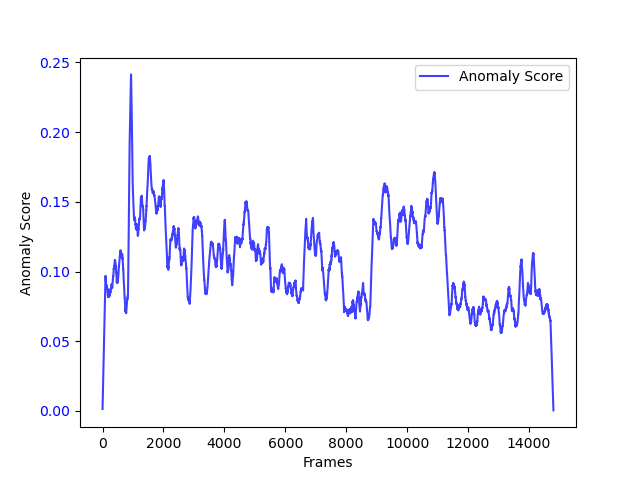}
        \caption{Mean.}
    \end{subfigure}%
    \begin{subfigure}[t]{0.5\linewidth}
        \centering
        \includegraphics[width=\linewidth]{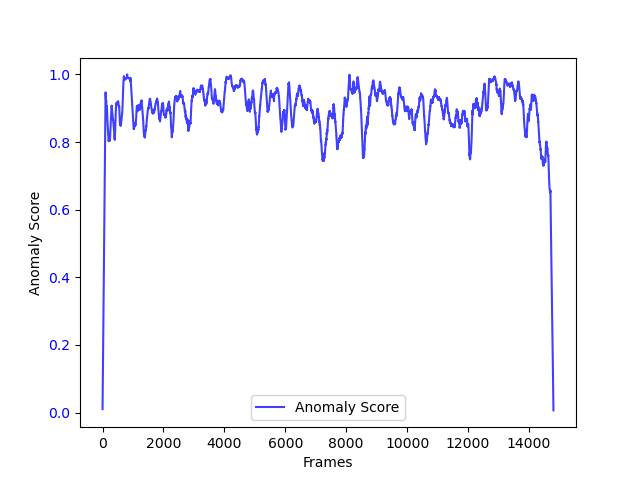}
        \caption{Non-zero mean.}
    \end{subfigure}
    \caption[Aggregation Functions on Noise Data]{Aggregation function performance on noisy data.}
    \label{fig:aggr_noisy}
\end{figure}
\begin{figure}[htb]
    \centering
    Clean data
    \begin{subfigure}[t]{0.5\linewidth}
        \centering
        \includegraphics[width=\linewidth]{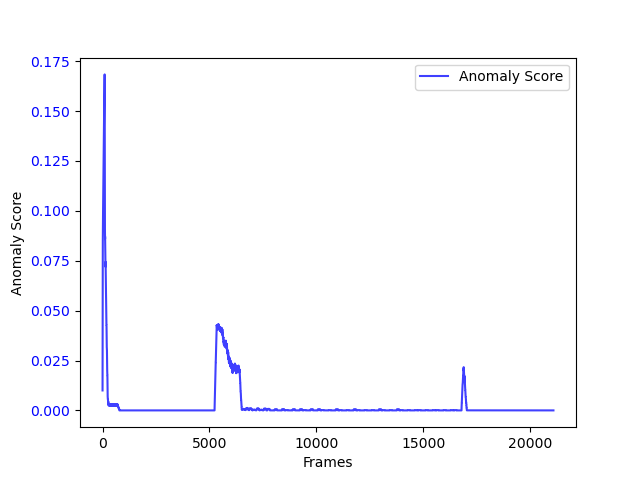}
        \caption{Mean.}

    \end{subfigure}%
    \begin{subfigure}[t]{0.5\linewidth}
        \centering
        \includegraphics[width=\linewidth]{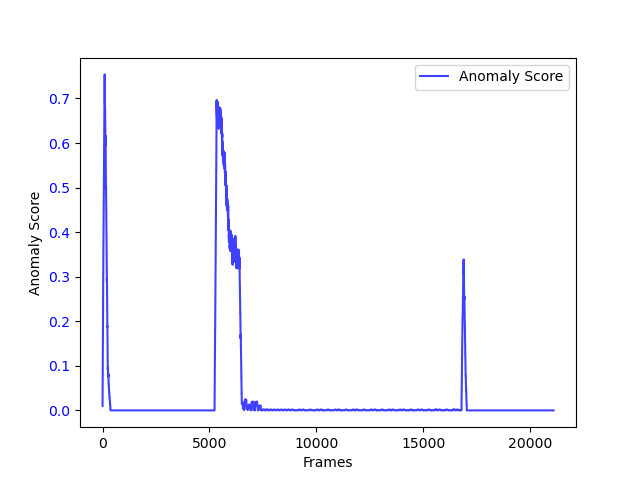}
        \caption{Non-zero mean.}
    \end{subfigure}
    \caption[Aggregation Functions on Clean Data]{Aggregation functions performance on clean data.}
    \label{fig:aggr_clean}
\end{figure}
\autoref{fig:aggr_noisy} illustrates the effect of an aggregation function for noisy data, where the non-zero mean is rendered useless due to the noise. On the other hand, \autoref{fig:aggr_clean} shows how the non-zero mean gives a clearer anomaly score when the data is clean. 

\textbf{Explainability. } Having the encoder output divided into a grid has the added benefit of introducing explainability into the model. By using Grid \gls*{htm} it is now possible to determine where in the input an anomaly has occurred by simply observing which cell has a high anomaly score.
It is also possible to estimate the number of predictions for each cell which can be used as a measure of certainty, where fewer predictions means higher certainty. Making it possible to measure certainty per cell creates a new source of information which can be used for explainability or robustness purposes.

\textbf{Flexibility and Performance. }
In addition, it is also possible to configure the \gls*{sp} and the \gls*{tm} in each cell independently, giving the architecture increased flexibility and to use a non-uniform grid, meaning that some cells can have different sizes. Last but not least, dividing the frame into smaller cells makes enables it to run each cell in parallel for increased performance.

\textbf{Reviewing Encoder Rules. }
A potential challenge with the grid approach is that the rules for creating a good encoder, may not be respected and therefore should be reviewed:
\begin{itemize}
    \item \textbf{Semantically similar data should result in SDRs with overlapping active bits}. In this example, a car at one position will produce an SDR with a high amount of overlapping bits as another car at a similar position in the input image.
    \item \textbf{The same input should always produce the same SDR}. The segmentation model produces a deterministic output given the same input.
    \item \textbf{The output must have the same dimensionality (total number of bits) for all inputs}. The segmentation model output has a fixed dimensionality.
    \item \textbf{The output should have similar sparsity (similar number of one-bits) for all inputs and have enough one-bits to handle noise and subsampling}. The segmentation model does not respect this. An example is that there can be no cars (zero active bits), one car ($n$ active bits), or two cars ($2n$ active bits), and that this will fluctuate over time.
\end{itemize}
The solution for the last rule is two-fold, and  consists of imposing a soft upper bound and a hard lower bound for the number of active pixels within a cell. The purpose is to lower the variation of number of active pixels, while also containing enough semantic information for the \gls*{htm} to work:
\begin{itemize}
    \item Pick a cell size so that the distribution of number of active pixels is as tight as possible, while containing enough semantic information and also being small enough so that the desired invariance is achieved. The cell size acts as a soft upper bound for the possible number of active pixels.
    \item Create a pattern representing emptiness, where the number of active bits is similar to what can be expected on average when there are cars inside a cell. This acts as a hard lower bound for the number of active pixels.
\end{itemize}
There could be situations where a few pixels are active within a cell, which could happen when a car has just entered a cell, but this is acceptable as long as it does not affect the distribution too much. If it does affect the distribution, which can be the case with noisy data, then an improvement would be to add a minimum sparsity requirement before a cell is considered not empty, e.g. less than 5 active pixels means that the cell is empty.  In the following example, the number of active pixels within a cell centered in the video was used to build the distributions seen in \autoref{fig:num_active_pixels_dist}:
\begin{figure}[htb]
    \centering
    \begin{subfigure}[t]{0.49\linewidth}
        \centering
        \includegraphics[width=\linewidth]{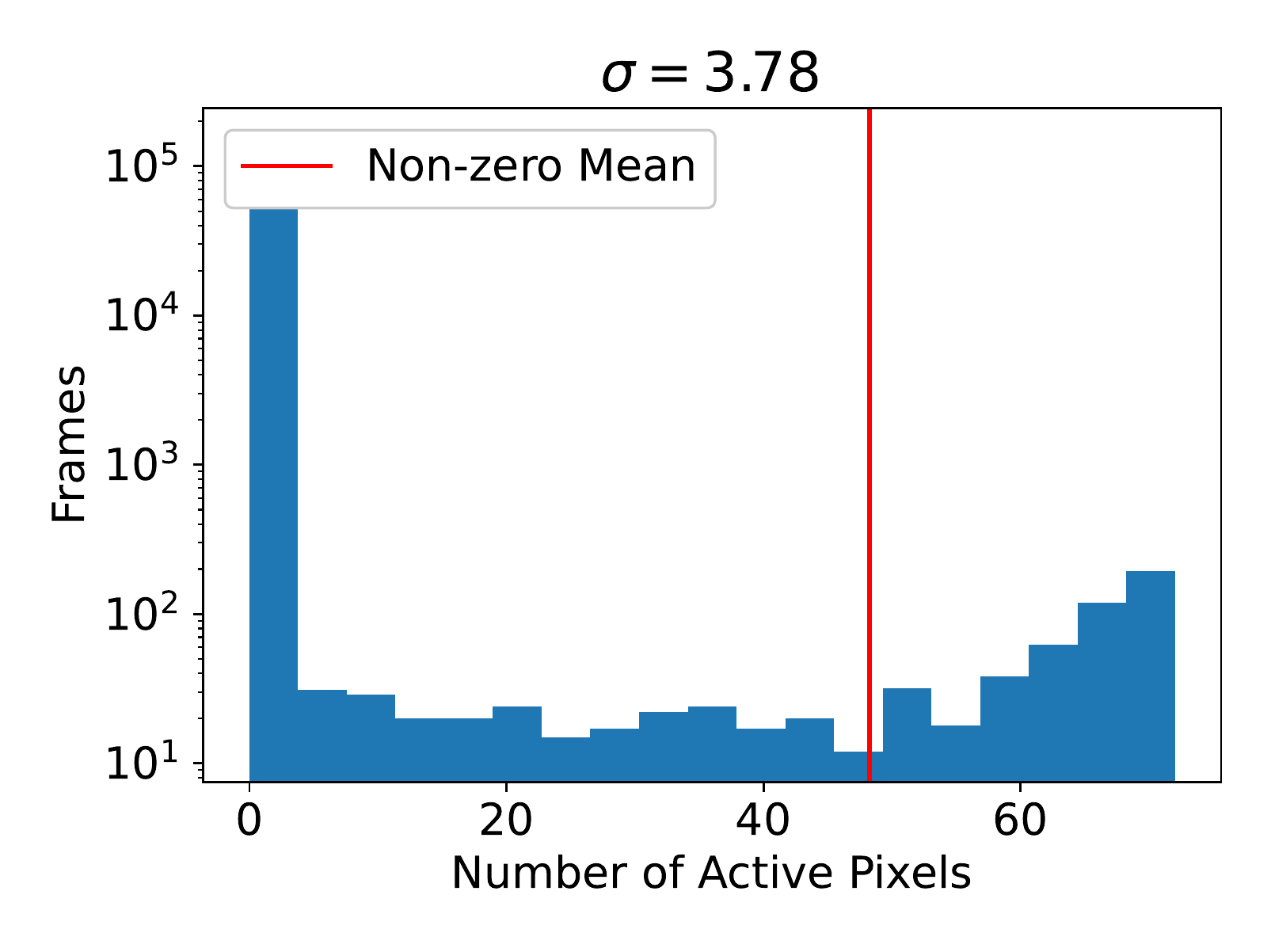}
        \caption{Without empty pattern.}
    \end{subfigure}%
    \begin{subfigure}[t]{0.49\linewidth}
        \centering
        \includegraphics[width=\linewidth]{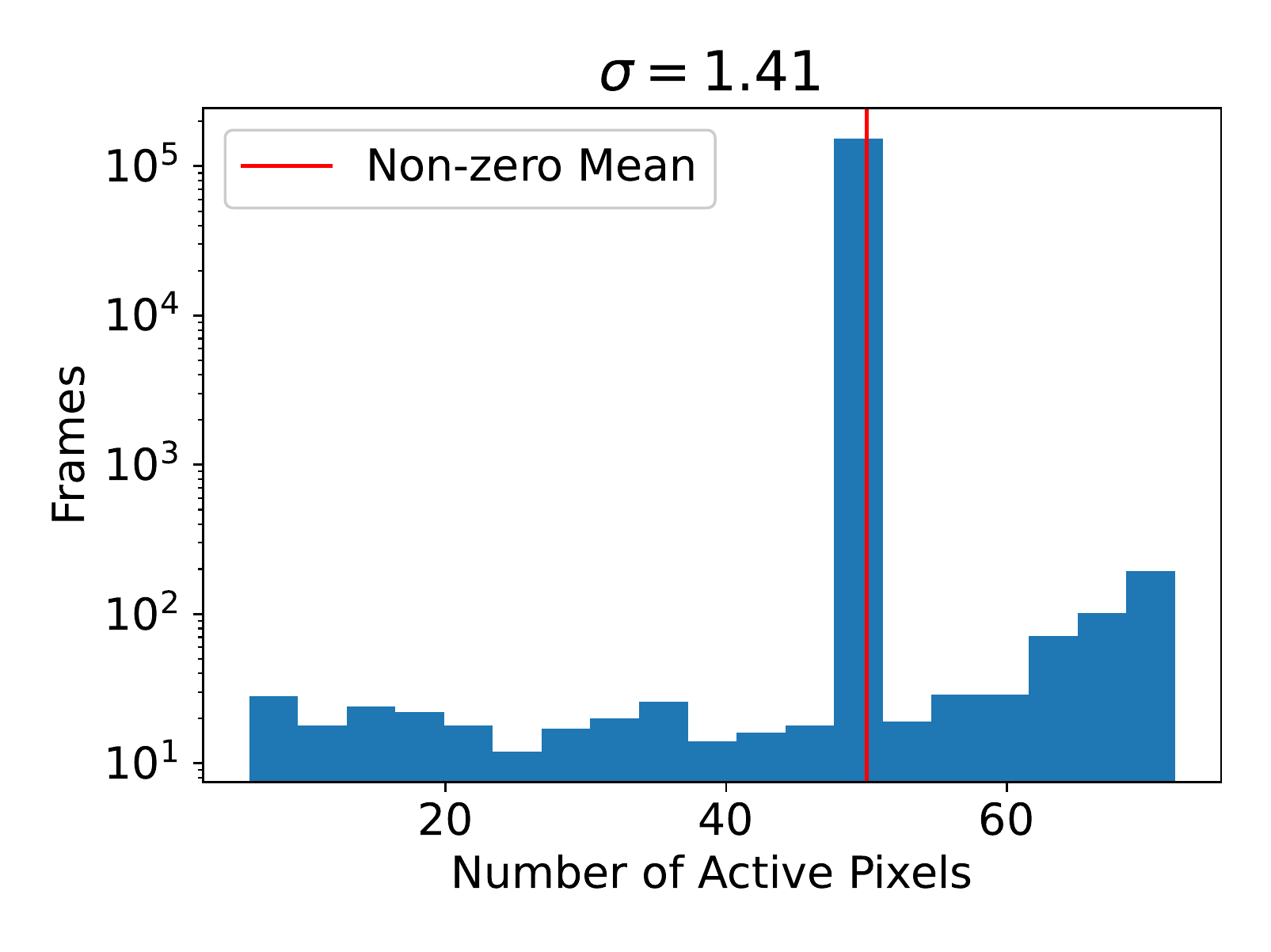}
        \caption{With empty pattern and a minimum sparsity requirement of $5$.}
    \end{subfigure}
    \caption[Distribution of Active Pixels]{Distribution of number of active pixels within a cell of size $12\times 12$.}
    \label{fig:num_active_pixels_dist}
\end{figure}

With a carefully selected empty pattern sparsity, the standard deviation of active pixels was lowered from $\mathbf{3.78}$ to $\mathbf{1.41}$. It is possible to automate this process by developing an algorithm which finds the optimal cell size and empty pattern sparsity which causes the least variation of number of active pixels per cell. This algorithm would run as a part of the calibration process.\par
\begin{figure}[htb]
    \centering
    \includegraphics[width=0.7\linewidth]{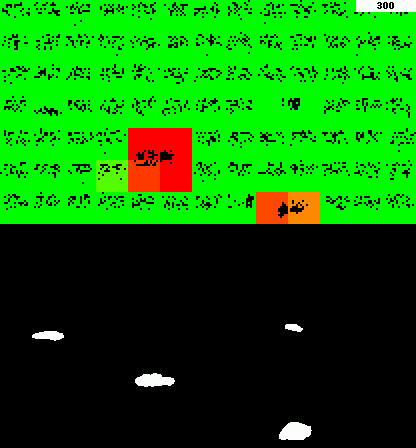}
    \caption[Example Grid HTM Output]{Example Grid \gls*{htm} output and the corresponding input. The color represents the anomaly score for each of the cells, where red means high anomaly score and green means zero anomaly score. Two of the cars are marked as anomalous because they are moving, which is something Grid \gls*{htm} has not seen before during its 300 frame (top right) long lifetime.}
    \label{fig:gridhtm_output}
\end{figure}

The visual output resulting from these changes, which is an equally important output as the aggregated anomaly score, can be seen in \autoref{fig:gridhtm_output} (for each cell red means higher anomaly score, green lower anomaly score).
Since there are now cells that are observing an empty pattern for a lot of the time in sparse data, boosting is recommended to be turned off, otherwise the \gls*{sp} output for the empty cells would change back and forth in order to adjust the active duty cycle.

\begin{figure}[htb]
    \centering
    \includegraphics[width=0.5\linewidth]{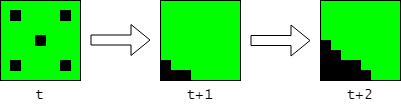}
    \caption[Stabilizing Anomaly Output Visualization 2]{The anomaly score is ignored (set to 0) for the frame in which the cell changes state from empty to not empty.}
    \label{fig:empty_to_notempty_fixed}
\end{figure}

\textbf{Stabilizing Anomaly Output. }
\begin{figure}[htb]
    \centering
    \includegraphics[width=0.5\linewidth]{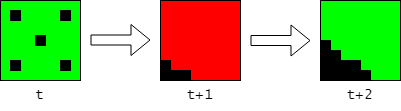}
    \caption[Stabilizing Anomaly Output Visualization 1]{High anomaly score when an empty cell (represented with an empty pattern with a sparsity value of 5) changes to being not empty, as something enters the cell.}
    \label{fig:empty_to_notempty}
\end{figure}
Another issue with the grid based approach is when a car first comes into a cell. The \gls*{tm} in that cell has no way of knowing that a car is about to enter, since it does not see outside its own cell, and therefore the first frame that a car enters a cell will cause a high anomaly output. This is illustrated in \autoref{fig:empty_to_notempty} where it can be observed that this effect causes the anomaly output to needlessly fluctuate. The band-aid solution is to ignore the anomaly score for the frame during which the cell goes from being empty to being not empty, which is illustrated in \autoref{fig:empty_to_notempty_fixed}.
A more proper solution could be to allow the \gls*{tm} to grow synapses to the TMs in the neighboring cells, but this is not documented in any research papers and might also hinder invariance.

\textbf{Multistep Temporal Patterns. }
Since the \gls*{tm} can only grow segments to cells that were active in the previous timestep, it will struggle to learn temporal patterns across multiple timesteps. This is especially evident in high framerate videos, where an object in motion has a similar representation at timestep $t$ and $t+1$, as an object standing still. 


This could cause situations where an object that is supposed to be moving, suddenly stands still, yet the \gls*{tm} will not mark it as an anomaly due to it being stuck in a contextual loop. 
A contextual loop is when one of the predictions at $t$ becomes true at $t+1$, and then one of the predictions at $t+1$ is almost identical to the state at $t$, which becomes true if the object is not moving, causing the \gls*{tm} to enter the same state that it was in at $t$.
A solution is to concatenate the past $n$ \gls*{sp} outputs as input into the TM, which is made possible by keeping a buffer of past \gls*{sp} outputs and shifting its contents out as new \gls*{sp} outputs are inserted. 
This follows the core idea behind encoding time in addition to the data, which makes time act as a contextual anchor. However, in this case there are no timestamps that are suitable to be used as contextual anchors, so as a replacement, the past observations are encoded instead.
\par
Concatenating past observations together will force the \gls*{tm} input, for when an object is in motion and when an object is still, to be unique. High framerate videos can benefit the most from this, and the effect will be more pronounced for higher values of $n$.
\par
A potential side effect of introducing temporal patterns, is that because the TM is now exposed to multiple frames at once, it will be more tolerant to temporal noise. An example of temporal noise is when an object disappears for a single frame due to falling below the classification threshold of the deep learning segmentation model encoder. The reason for the noise tolerance is that instead of the temporal noise making up the entire input for the TM, it now only makes up $\frac{1}{n}$ of the TM input.

\textbf{Use Cases. }
The most intuitive use case is to use Grid \gls*{htm} for semi-active surveillance, where personnel only have to look at segments containing anomalies, leading to drastically increased efficiency. One example is making it possible to have an entire city be monitored by a few people. This is made possible by making it so that people only have to look at segments that Grid \gls*{htm} has found anomalous, which is what drastically lowers the manpower requirement for active monitoring of the entire city.

\section{Experimental details and results}
As stated earlier, one of the use cases of Grid \gls*{htm} is anomaly detection in surveillance, and we using a video from the VIRAT~\cite{VIRAT} video dataset with long duration and a stationary camera, we demonstrate our system. The video consists of technical anomalies in the form of several segments with sudden frame skips in between. There is also a synthetic anomaly introduced in the form of a frame repeat lasting a couple of seconds, essentially "freezing" time, in order to test whether Grid \gls*{htm} is able to understand how objects should be moving in time.
\par
In this experiment, a segmentation model which can extract classes into their respective SDRs is employed. Meaning that there could be an SDR for cars and an SDR for persons, that are then concatenated before being fed into the system. The segmentation model used is PointRend~\cite{pointrend} with a ResNet101~\cite{resnet} backbone, pretrained on ImageNet~\cite{imagenet}, and implemented using PixelLib~\cite{pixellib}. For the sake of simplicity, this experiment will focus only on the segmentation of cars. While on the topic of segmentation, it is important to mention that the segmentation model is not perfect and that there are cases where objects are misclassified as well as cases where cars repeatedly go above and below the confidence threshold.

\begin{figure}[htb]
    \centering
    \includegraphics[width=\linewidth]{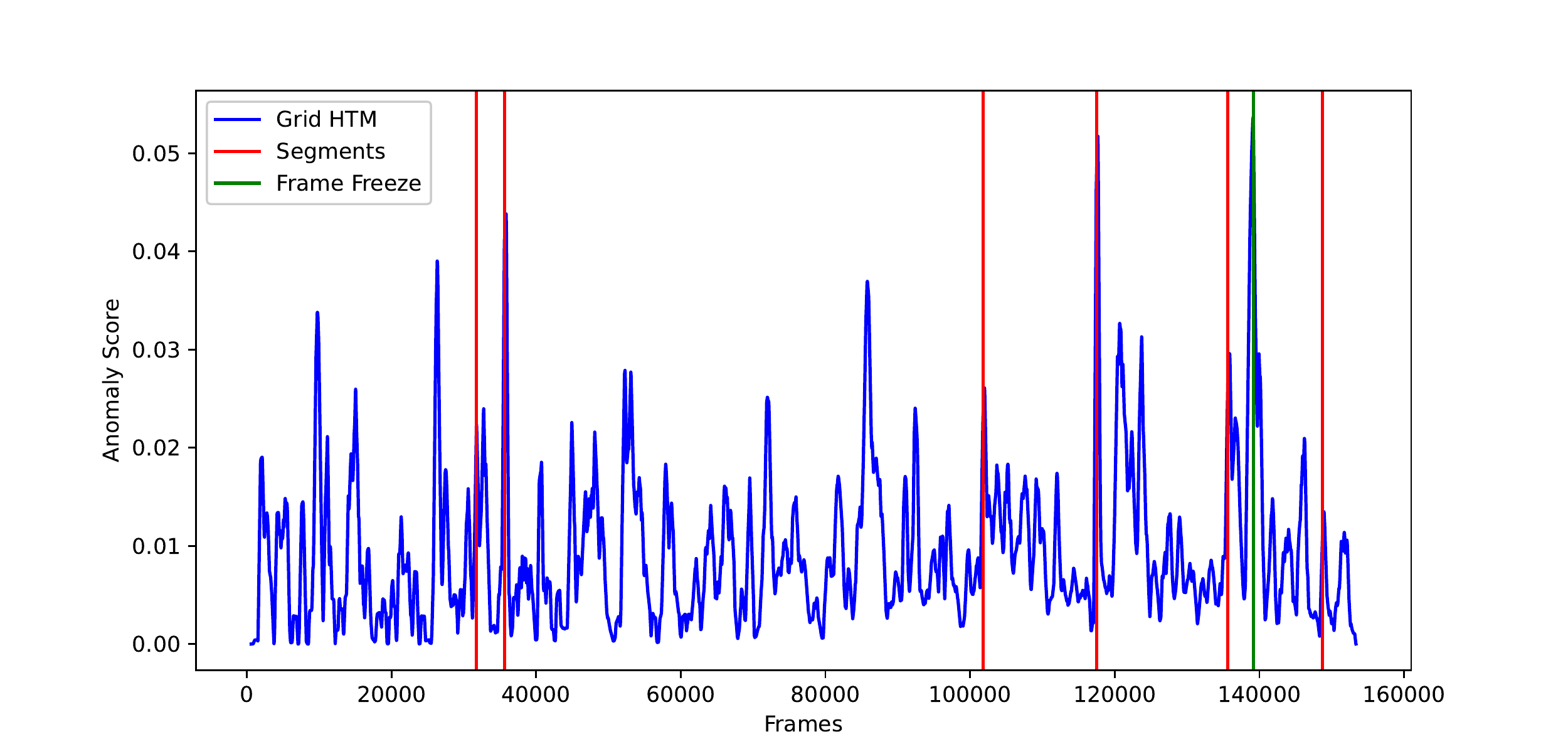}
    \caption[Grid HTM Anomaly Score Output]{Anomaly score output from Grid HTM.}
    \label{fig:surveillance_results}
\end{figure}
We can see in \autoref{fig:surveillance_results} that Grid \gls*{htm} is detecting when segments begin and end, however it is not possible to use a threshold value to isolate them, and they also have vastly different anomaly scores compared to each other. This is due to the way the aggregation function works, which means that the anomaly output is dependent on the physical size of the anomaly. It should also be noted that a moving average ($n=200$) was applied to smooth out the anomaly score output, otherwise the graph would be too noisy.

\begin{figure}[htb]
    \centering
    \includegraphics[width=0.7\linewidth]{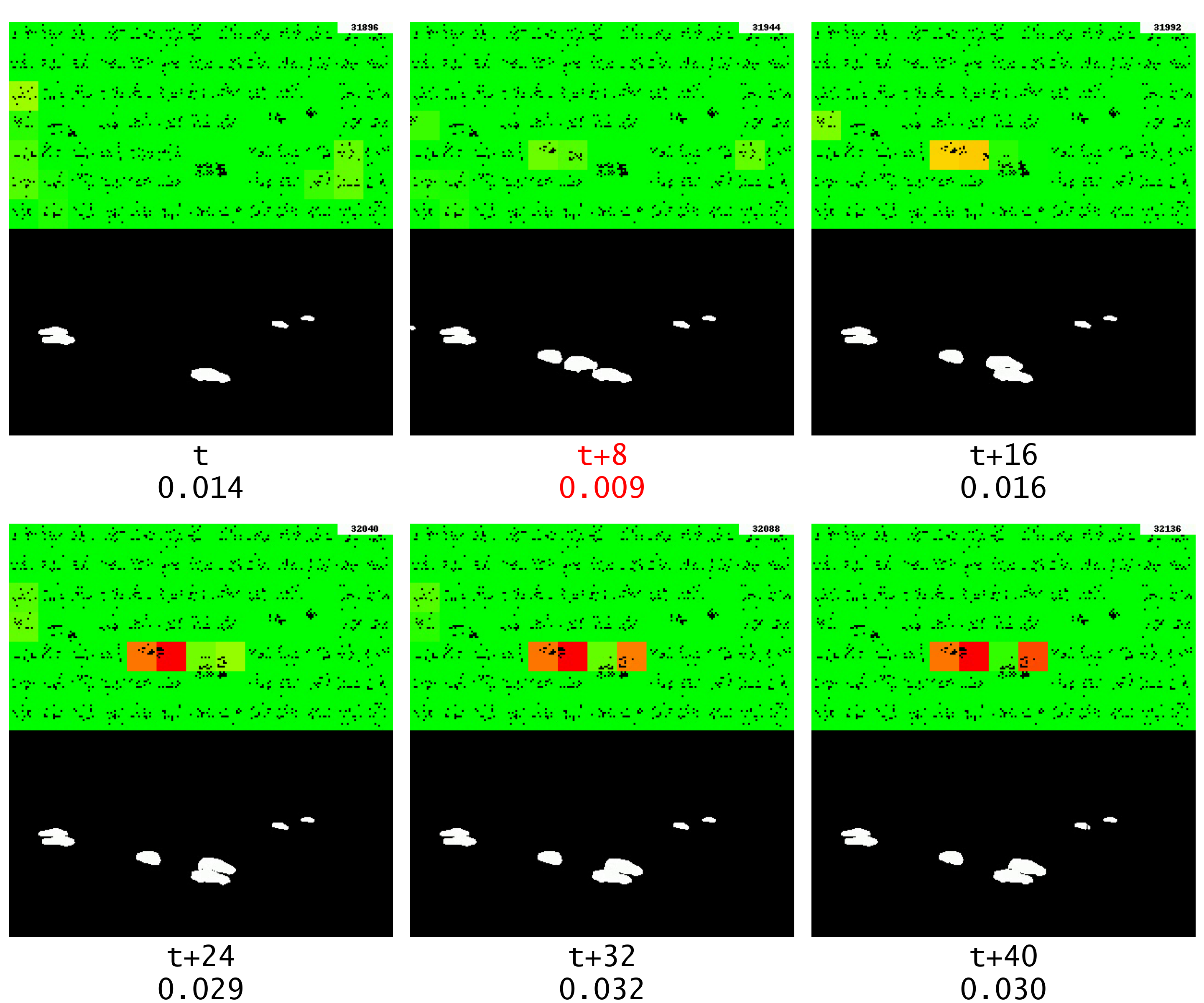}
    \caption[Segment Anomaly]{The first segment anomaly, which is marked with red text,  and the corresponding changes detected by Grid HTM. The numbers beneath each frame represent the relative frame number and the current anomaly score respectively.}
    \label{fig:surveillance_segment}
\end{figure}

With the aggregation functions presented in this paper in mind, it is safe to conclude that looking at the anomaly score output is meaningless for complex data such as a surveillance video. This however does not mean that Grid \gls*{htm} is completely useless, and this can be observed by looking at the visual output of Grid HTM. The visual output during which the first segment anomaly occurs can be seen in \autoref{fig:surveillance_segment}. Here, it is observed that Grid \gls*{htm} correctly marks the sudden change of cars when the current segment ends and a new segment begins.

\begin{figure}[htb]
    \centering
    \includegraphics[width=0.7\linewidth]{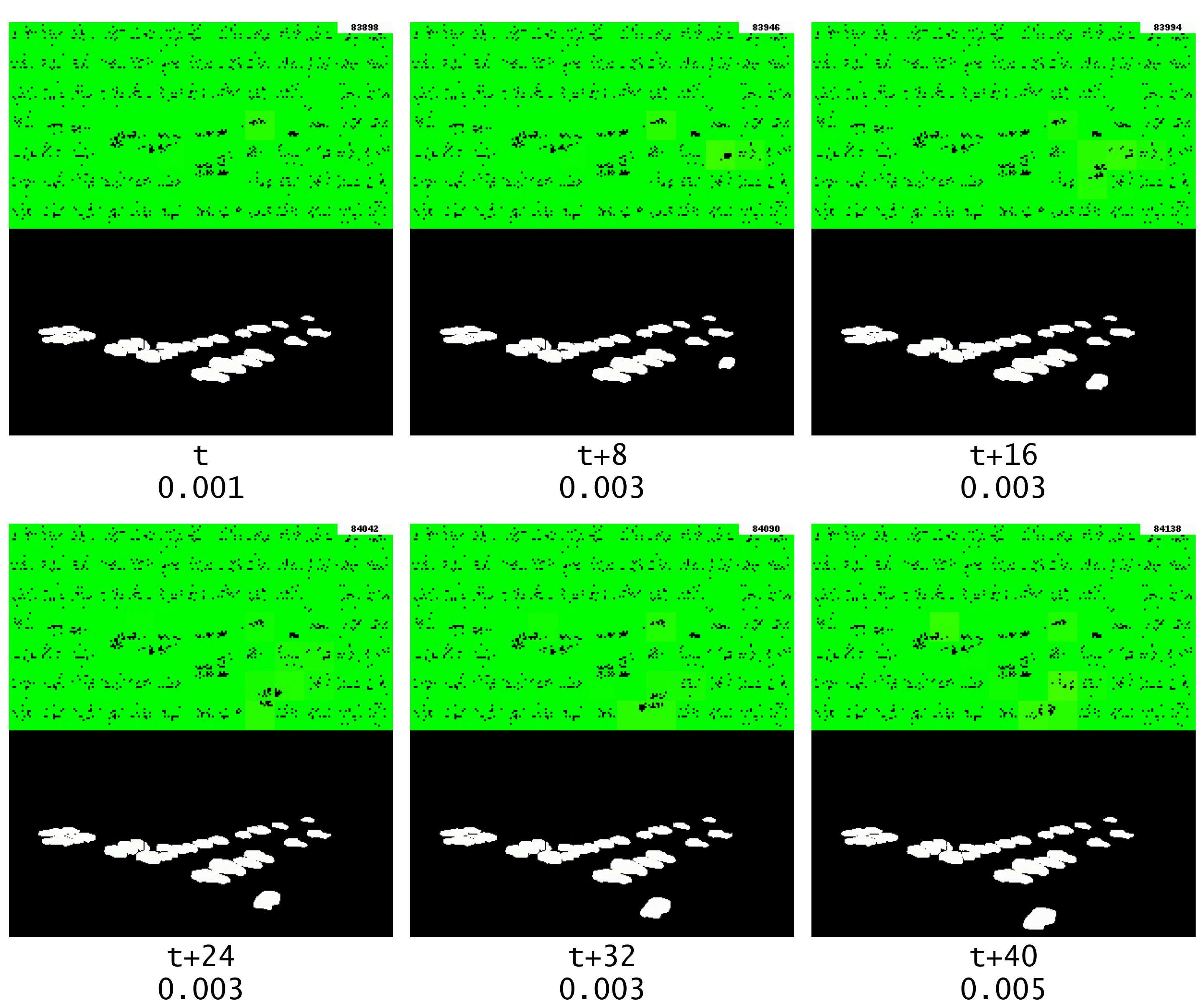}
    \caption[Car Driving Along Main Road]{Visual output when a car is driving along a road.}
    \label{fig:surveillance_road_1}
\end{figure}

In the original video, there is a road on which cars regularly drive. By observing the visual output, it becomes evident that after some time Grid \gls*{htm} has mostly learned that behavior and does not report those moving cars as anomalies. This is shown in \autoref{fig:surveillance_road_1}.

To prove that Grid \gls*{htm} has learned that cars on the road should be moving, it is possible to look at the visual output during the period when the video is repeating the same frame and observe if the architecture marks the cars standing still on the road as anomalies.
\begin{figure}[htb]
    \centering
    \includegraphics[width=0.7\linewidth]{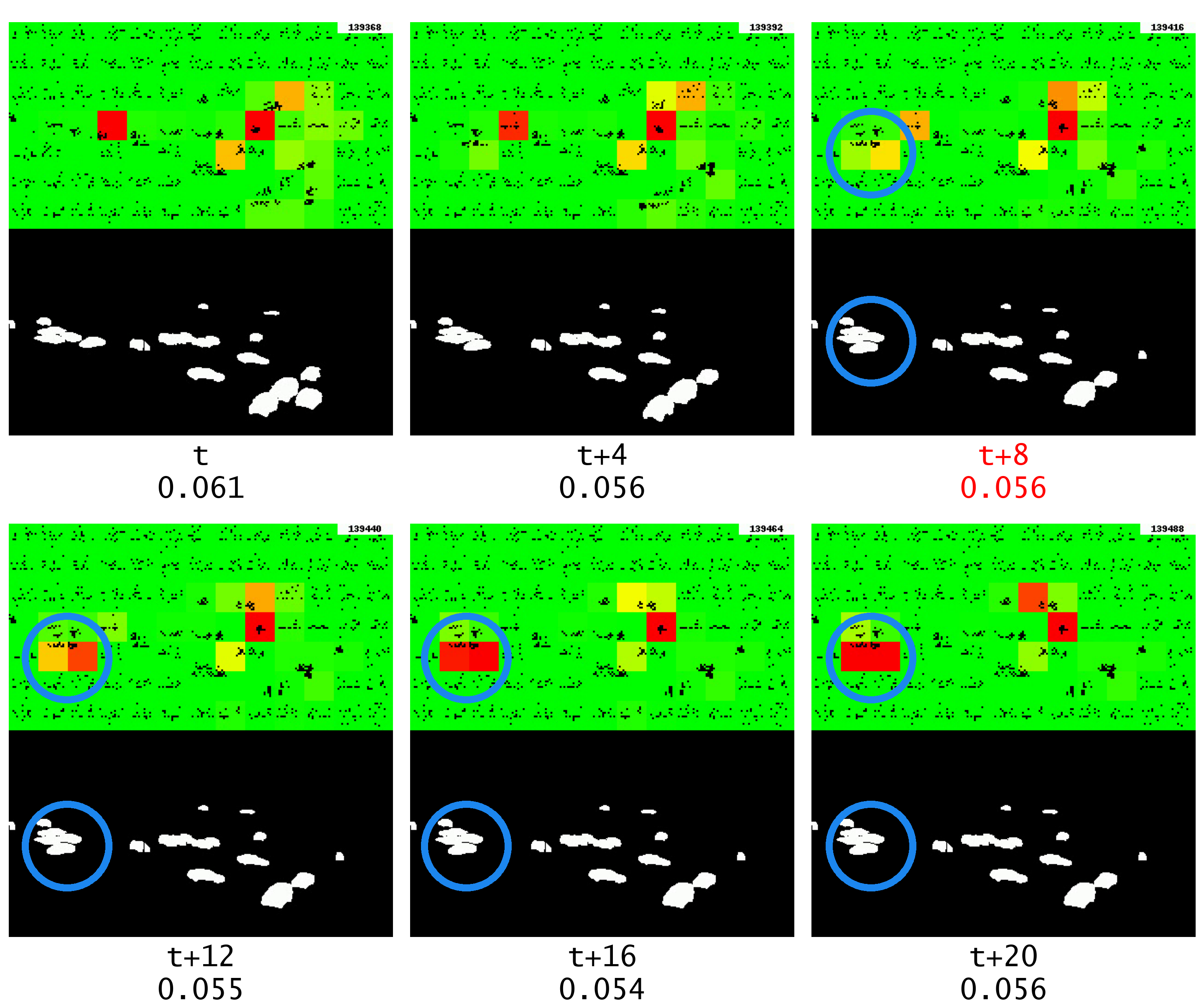}
    \caption[Frame Repeat Anomaly]{Anomaly output during the repeating frame, the start of the frame repeat is marked with red text. The blue circle highlights the object of interest.}
    \label{fig:surveillance_freeze_1}
\end{figure}
It can be observed in \autoref{fig:surveillance_freeze_1} that the cars along the main road are not marked as anomalies, but this could be attributed to the fact that there is a crossing there and that cars periodically have to stop at that point to let pedestrians cross.
\par
On the other hand, when looking at the anomaly marked with a blue circle, the car on the road in the parking lot is marked as an anomaly that increases in severity as the time goes on during the frame repeat. The reason why that car causes an anomaly is because, unlike the cars on the main road, a car is rarely observed as standing still at that position.
To prove that the anomaly was actually directly caused by the repeating frame, and not just due to repeating the anomaly in time, it should be compared to the anomaly output if there was no repeating frame.
\begin{figure}[htb]
    \centering
    \includegraphics[width=0.7\linewidth]{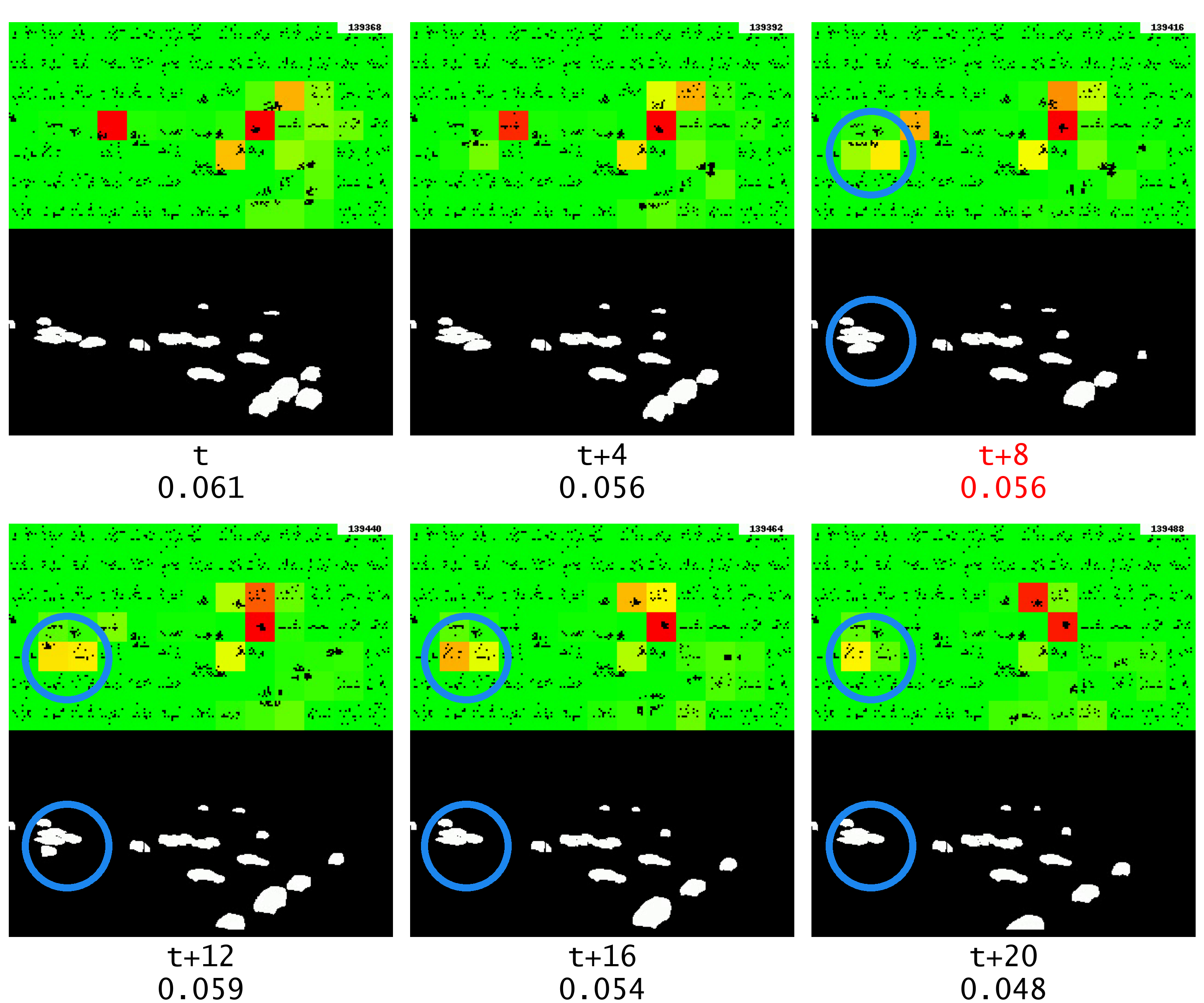}
    \caption[No Frame Repeat Anomaly]{Anomaly output when there is no frame repeating, where it should have repeated is marked in red. The blue circle highlights the object of interest.}
    \label{fig:surveillance_freeze_no_freeze}
\end{figure}
It can be observed in \autoref{fig:surveillance_freeze_no_freeze} that the anomaly output is minor compared to when there was a repeating frame, proving that the anomaly was indeed a product of the repeating frame and that Grid \gls*{htm} was able to learn how objects should be moving in time.
\par
\begin{figure}[htb]
    \centering
    \includegraphics[width=0.7\linewidth]{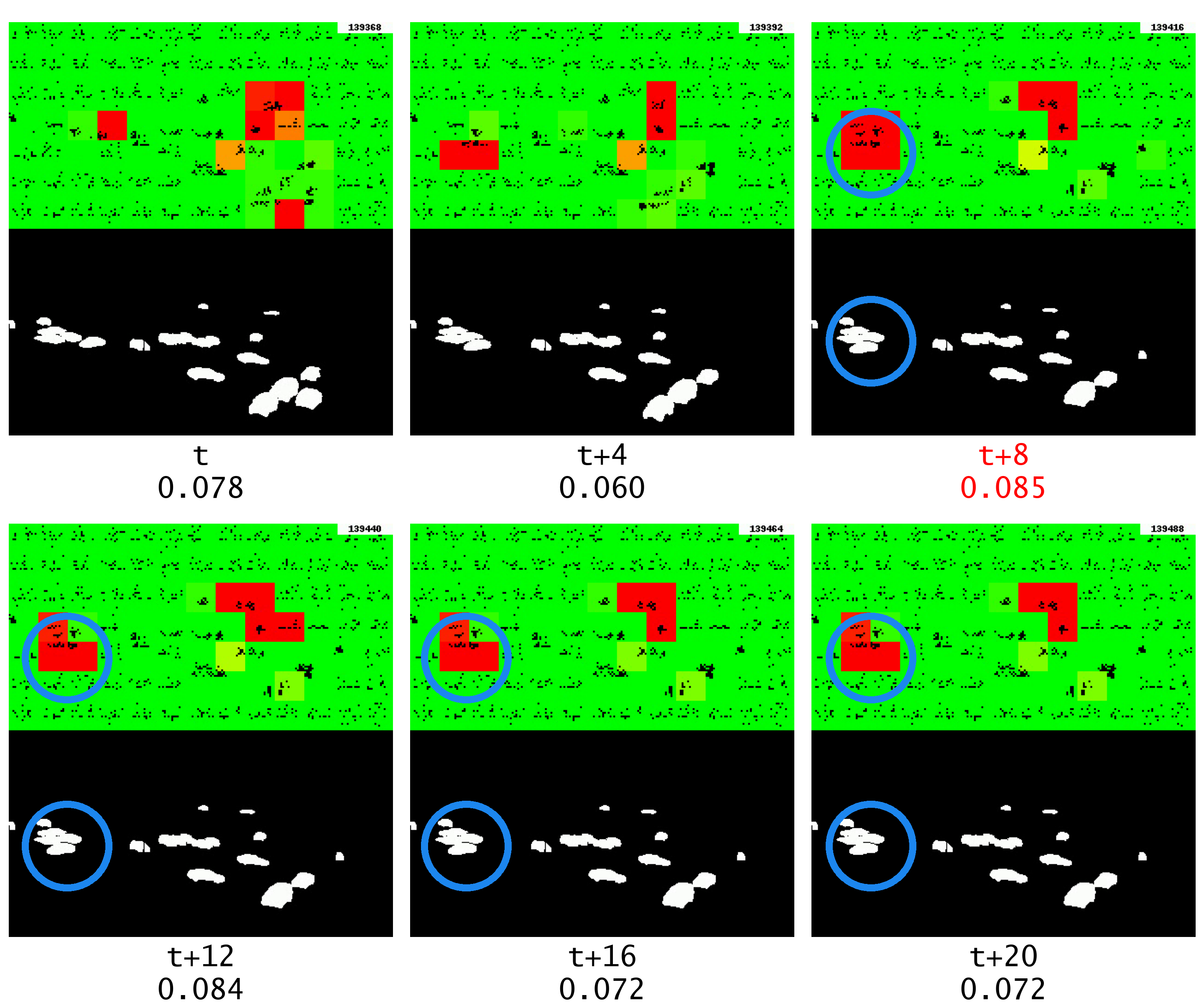}
    \caption[Frame Repeat No MTP Anomaly]{Anomaly output during the repeating frame, the start of the frame repeat is marked with red text. The blue circle highlights the object of interest. This time without multistep temporal patterns.}
    \label{fig:surveillance_freeze_no_mtp}
\end{figure}

Finally, it is interesting to look at how Grid \gls*{htm} handles the repeating frames without multistep temporal patterns, which is shown in \autoref{fig:surveillance_freeze_no_mtp}.
Unfortunately, simply disabling multistep temporal patterns without adjusting the other \gls*{tm} parameters causes the same car to be marked as an anomaly before and during the frame repeat. In fact, as previously mentioned, disabling multistep temporal patterns causes Grid \gls*{htm} to be less noise tolerant which causes a lot more anomalies to be wrongly detected. This is evident in \autoref{fig:surveillance_freeze_no_mtp}, where a higher number of severe anomalies can be observed compared to previous examples. This also highlights how sensitive \gls*{htm} can be regarding parameters. The working code for Grid HTM and the parameters for the experiments conducted in for this paper can be found on \href{https://github.com/vladim0105/GridHTM}{GitHub}\footnote{\url{https://github.com/vladim0105/GridHTM}}.

\section{Conclusion}
We presented a novel method to perform anomaly detection in videos. Experiments showed that the proposed Grid HTM can be used for unsupervised anomaly detection in complex videos such as surveillance footage. One of the most important future work would be to create a dataset with videos that are several days long and contain anomalies such as car accidents, jaywalking, and other similar anomalous behaviors.
For Grid HTM, more time can be spent exploring other aggregation functions so that the aggregated anomaly score can be used more efficiently. 
Additionally, it would be a big benefit to create an algorithm which can decide the parameters for each cell during the calibration phase. It is also possible to improve explainability and robustness by implementing a measure of certainty for each cell.
\par
Finally, experiments should be performed to validate the possibility of having the TM in each cell grow synapses to neighboring cells in order to solve the issue with unstable anomaly output.



\appendix
\balance
\printbibliography

@misc{us_video_stats,
  title        = {{U.S. Video Surveillance Market by Component (Solution, Service, and Connectivity Technology), Application (Commercial, Military \& Defense, Infrastructure, Residential, and Others), and Customer Type (B2B and B2C): Opportunity Analysis and Industry Forecast, 2020–2027}},
  author       = {Divyanshi Tewari},
  year         = {2019},
  month        = {3},
  howpublished = {Online},
  url          = {https://www.alliedmarketresearch.com/us-video-surveillance-market-A06741}
}

@unpublished{BAMI,
  title  = {{Biological and Machine Intelligence (BAMI)}},
  author = {Hawkins, J. and Ahmad, S. and Purdy, S. and Lavin, A.},
  note   = {Initial online release 0.4},
  url    = {https://numenta.com/resources/biological-and-machine-intelligence/},
  year   = {2016}
}

@article{anomaly_detection,
  title     = {{Deep Learning for Anomaly Detection}},
  volume    = {54},
  issn      = {1557-7341},
  url       = {http://dx.doi.org/10.1145/3439950},
  doi       = {10.1145/3439950},
  number    = {2},
  journal   = {ACM Computing Surveys},
  publisher = {Association for Computing Machinery (ACM)},
  author    = {Pang, Guansong and Shen, Chunhua and Cao, Longbing and Hengel, Anton Van Den},
  year      = {2021},
  month     = {4},
  pages     = {1-38}
}

@misc{anomalyvideo,
  doi          = {10.48550/ARXIV.2004.00222},
  url          = {https://arxiv.org/abs/2004.00222},
  author       = {Zhu, Sijie and Chen, Chen and Sultani, Waqas},
  title        = {{Video Anomaly Detection for Smart Surveillance}},
  publisher    = {arXiv},
  year         = {2020},
  copyright    = {arXiv.org perpetual, non-exclusive license},
  howpublished = {Online}
}

@article{noise1,
  title    = {{Dealing with Noise Problem in Machine Learning Data-sets: A Systematic Review}},
  journal  = {Procedia Computer Science},
  volume   = {161},
  pages    = {466-474},
  year     = {2019},
  note     = {The Fifth Information Systems International Conference, 23-24 July 2019, Surabaya, Indonesia},
  issn     = {1877-0509},
  doi      = {https://doi.org/10.1016/j.procs.2019.11.146},
  url      = {https://www.sciencedirect.com/science/article/pii/S1877050919318575},
  author   = {Shivani Gupta and Atul Gupta}
}

@misc{noise2,
  doi          = {10.48550/ARXIV.1903.12261},
  url          = {https://arxiv.org/abs/1903.12261},
  author       = {Hendrycks, Dan and Dietterich, Thomas},
  title        = {{Benchmarking Neural Network Robustness to Common Corruptions and Perturbations}},
  publisher    = {arXiv},
  year         = {2019},
  howpublished = {Online},
  copyright    = {arXiv.org perpetual, non-exclusive license}
}

@misc{deeplearning_dataset,
  doi          = {10.48550/ARXIV.1707.02968},
  url          = {https://arxiv.org/abs/1707.02968},
  author       = {Sun, Chen and Shrivastava, Abhinav and Singh, Saurabh and Gupta, Abhinav},
  title        = {{Revisiting Unreasonable Effectiveness of Data in Deep Learning Era}},
  publisher    = {arXiv},
  year         = {2017},
  copyright    = {arXiv.org perpetual, non-exclusive license},
  howpublished = {Online}
}

@misc{deeplearning_ood_generalization_survey,
  doi       = {10.48550/ARXIV.2108.13624},
  url       = {https://arxiv.org/abs/2108.13624},
  author    = {Shen, Zheyan and Liu, Jiashuo and He, Yue and Zhang, Xingxuan and Xu, Renzhe and Yu, Han and Cui, Peng},
  title     = {{Towards Out-Of-Distribution Generalization: A Survey}},
  publisher = {arXiv},
  year      = {2021},
  copyright = {Creative Commons Attribution Non Commercial Share Alike 4.0 International}
}

@misc{deeplearning_ood,
  doi       = {10.48550/ARXIV.2011.03395},
  url       = {https://arxiv.org/abs/2011.03395},
  author    = {D'Amour, Alexander and Heller, Katherine and Moldovan, Dan and Adlam, Ben and Alipanahi, Babak and Beutel, Alex and Chen, Christina and Deaton, Jonathan and Eisenstein, Jacob and Hoffman, Matthew D. and Hormozdiari, Farhad and Houlsby, Neil and Hou, Shaobo and Jerfel, Ghassen and Karthikesalingam, Alan and Lucic, Mario and Ma, Yian and McLean, Cory and Mincu, Diana and Mitani, Akinori and Montanari, Andrea and Nado, Zachary and Natarajan, Vivek and Nielson, Christopher and Osborne, Thomas F. and Raman, Rajiv and Ramasamy, Kim and Sayres, Rory and Schrouff, Jessica and Seneviratne, Martin and Sequeira, Shannon and Suresh, Harini and Veitch, Victor and Vladymyrov, Max and Wang, Xuezhi and Webster, Kellie and Yadlowsky, Steve and Yun, Taedong and Zhai, Xiaohua and Sculley, D.},
  title     = {{Underspecification Presents Challenges for Credibility in Modern Machine Learning}},
  publisher = {arXiv},
  year      = {2020},
  copyright = {arXiv.org perpetual, non-exclusive license}
}

@article{XAI,
  title   = {{Explainable Artificial Intelligence (XAI): Concepts, taxonomies, opportunities and challenges toward responsible AI}},
  journal = {Information Fusion},
  volume  = {58},
  pages   = {82-115},
  year    = {2020},
  issn    = {1566-2535},
  doi     = {https://doi.org/10.1016/j.inffus.2019.12.012},
  url     = {https://www.sciencedirect.com/science/article/pii/S1566253519308103},
  author  = {Alejandro {Barredo Arrieta} and Natalia Díaz-Rodríguez and Javier {Del Ser} and Adrien Bennetot and Siham Tabik and Alberto Barbado and Salvador Garcia and Sergio Gil-Lopez and Daniel Molina and Richard Benjamins and Raja Chatila and Francisco Herrera}
}

@inproceedings{CNN_HTM,
  author    = {Y. {Zou} and Y. {Shi} and Y. {Wang} and Y. {Shu} and Q. {Yuan} and Y. {Tian}},
  booktitle = {Proceedings of the 2018 IEEE International Conference on Multimedia and Expo (ICME)},
  title     = {{Hierarchical Temporal Memory Enhanced One-Shot Distance Learning for Action Recognition}},
  year      = {2018},
  volume    = {},
  number    = {},
  pages     = {1-6},
  doi       = {10.1109/ICME.2018.8486447},
  url       = {https://doi.org/10.1109/ICME.2018.8486447}
}

@misc{eyeencoder,
  author       = {David McDougall (ctrl-z-9000-times)},
  year         = {2019},
  month        = {9},
  day          = {20},
  url          = {https://github.com/htm-community/htm.core/issues/259#issuecomment-533333336},
  howpublished = {Online}
}

@inbook{MotionAnomalyDetection,
  author    = {Daylidyonok, Ilya and Frolenkova, Anastasiya and Panov, Aleksandr I.},
  editor    = {Samsonovich, Alexei V.},
  title     = {{Extended Hierarchical Temporal Memory for Motion Anomaly Detection}},
  booktitle = {Biologically Inspired Cognitive Architectures 2018},
  year      = {2019},
  publisher = {Springer International Publishing},
  address   = {Cham},
  pages     = {69-81},
  isbn      = {978-3-319-99316-4},
  doi       = {10.1007/978-3-319-99316-4_10},
  url       = {https://doi.org/10.1007/978-3-319-99316-4_10}
}

@article{AHMAD2017134,
  title    = {{Unsupervised real-time anomaly detection for streaming data}},
  journal  = {Neurocomputing},
  volume   = {262},
  pages    = {134-147},
  year     = {2017},
  note     = {Online Real-Time Learning Strategies for Data Streams},
  issn     = {0925-2312},
  doi      = {https://doi.org/10.1016/j.neucom.2017.04.070},
  url      = {http://www.sciencedirect.com/science/article/pii/S0925231217309864},
  author   = {Subutai Ahmad and Alexander Lavin and Scott Purdy and Zuha Agha},
}

@inproceedings{orb_detector,
  author    = {Rublee, Ethan and Rabaud, Vincent and Konolige, Kurt and Bradski, Gary},
  booktitle = {Proceedings of the 2011 International Conference on Computer Vision (ICCV)},
  title     = {{ORB: An efficient alternative to SIFT or SURF}},
  year      = {2011},
  volume    = {},
  number    = {},
  pages     = {2564-2571},
  doi       = {10.1109/ICCV.2011.6126544},
  url       = {https://doi.org/10.1109/ICCV.2011.6126544}
}

@inproceedings{VIRAT,
  author    = {Oh, Sangmin and Hoogs, Anthony and Perera, Amitha and Cuntoor, Naresh and Chen, Chia-Chih and Lee, Jong Taek and Mukherjee, Saurajit and Aggarwal, J. K. and Lee, Hyungtae and Davis, Larry and Swears, Eran and Wang, Xioyang and Ji, Qiang and Reddy, Kishore and Shah, Mubarak and Vondrick, Carl and Pirsiavash, Hamed and Ramanan, Deva and Yuen, Jenny and Torralba, Antonio and Song, Bi and Fong, Anesco and Roy-Chowdhury, Amit and Desai, Mita},
  booktitle = {Proceedings of the 2013 IEEE Conference on Computer Vision and Pattern Recognition (CVPR)},
  title     = {{A large-scale benchmark dataset for event recognition in surveillance video}},
  year      = {2011},
  volume    = {},
  number    = {},
  pages     = {3153-3160},
  doi       = {10.1109/CVPR.2011.5995586},
  url       = {https://doi.org/10.1109/CVPR.2011.5995586}
}

@misc{pointrend,
  doi          = {10.48550/ARXIV.1912.08193},
  url          = {https://arxiv.org/abs/1912.08193},
  author       = {Kirillov, Alexander and Wu, Yuxin and He, Kaiming and Girshick, Ross},
  title        = {{PointRend: Image Segmentation as Rendering}},
  publisher    = {arXiv},
  year         = {2019},
  howpublished = {Online},
  copyright    = {arXiv.org perpetual, non-exclusive license}
}

@inproceedings{resnet,
  author    = {He, Kaiming and Zhang, Xiangyu and Ren, Shaoqing and Sun, Jian},
  booktitle = {Proceedings of the 2016 IEEE Conference on Computer Vision and Pattern Recognition (CVPR)},
  title     = {{Deep Residual Learning for Image Recognition}},
  year      = {2016},
  volume    = {},
  number    = {},
  pages     = {770-778},
  doi       = {10.1109/CVPR.2016.90},
  url       = {https://doi.org/10.1109/CVPR.2016.90}
}

@inproceedings{imagenet,
  author    = {Deng, Jia and Dong, Wei and Socher, Richard and Li, Li-Jia and Kai Li and Li Fei-Fei},
  booktitle = {Proceedings of the 2009 IEEE Conference on Computer Vision and Pattern Recognition (CVPR)},
  title     = {{ImageNet: A large-scale hierarchical image database}},
  year      = {2009},
  volume    = {},
  number    = {},
  pages     = {248-255},
  doi       = {10.1109/CVPR.2009.5206848},
  url       = {https://doi.org/10.1109/CVPR.2009.5206848}
}

@misc{pixellib,
  title        = {{Simplifying Object Segmentation with PixelLib Library}},
  author       = {Ayoola Olafenwa},
  year         = {2021},
  howpublished = {Online},
  url          = {https://vixra.org/abs/2101.0122}
}
\end{document}